\documentclass[runningheads]{llncs}
\usepackage{graphicx}

\usepackage{lineno,hyperref}
\usepackage{url}
\usepackage{algorithm}
\usepackage{caption2}
\usepackage{subfigure}
\usepackage{float}
\usepackage{booktabs}
\usepackage{threeparttable}
\usepackage{multirow}
\usepackage{multicol}
\usepackage{arydshln}
\usepackage{amsfonts}
\usepackage{amsmath}
\usepackage{microtype}
\usepackage[mathscr]{eucal}
\usepackage{picinpar}
\usepackage{wrapfig}
\usepackage{amssymb}
\usepackage{marvosym}
\usepackage{ifsym}

\begin{document}
\title{Towards Both Accurate and Robust Neural Networks without Extra Data}
%
%
\author{Faqiang Liu\inst{1}\orcidID{0000-0002-2236-0539} \and \Letter
Rong Zhao\inst{1}\orcidID{0000-0002-2320-0326} }
\authorrunning{F. Liu et al.}
%
\institute{Center for Brain-Inspired Computing Research, Department of Precision Instrument, Tsinghua University, Beijing, China
\email{lfq18@mails.tsinghua.edu.cn, r\_zhao@mail.tsinghua.edu.cn}\\
}
\maketitle              
\begin{abstract}
Deep neural networks have achieved remarkable performance in various applications but are extremely vulnerable to adversarial perturbation. The most representative and promising methods that can enhance model robustness, such as adversarial training and its variants, substantially degrade model accuracy on benign samples, limiting practical utility. Although incorporating extra training data can alleviate the trade-off to a certain extent, it remains unsolved to achieve both robustness and accuracy under limited training data. Here, we demonstrate the feasibility of overcoming the trade-off, by developing an adversarial feature stacking (AFS) model, which combines multiple independent feature extractors with varied levels of robustness and accuracy. Theoretical analysis is further conducted, and general principles for the selection of basic feature extractors are provided. We evaluate the AFS model on CIFAR-10 and CIFAR-100 datasets with strong adaptive attack methods, significantly advancing the state-of-the-art in terms of the trade-off. The AFS model achieves a benign accuracy improvement of $\sim$6\% on CIFAR-10 and $\sim$10\% on CIFAR-100 with comparable or even stronger robustness than the state-of-the-art adversarial training methods. 

\keywords{adversarial robustness \and deep networks \and adversarial training}
\end{abstract}

\section{Introduction}
With the assistance of big data and powerful parallel computing platforms, deep neural networks (DNNs) \cite{lecun2015deep,pei2019towards,liu2021adversarial,li2019super} have achieved significant success in computer vision and natural language processing. However, DNNs are extremely vulnerable to adversarial perturbation, which is imperceptible by humans but can fool the state-of-the-art deep models to give wrong predictions \cite{szegedy2014intriguing}. The poor robustness of DNNs hinders applications of DNNs in security-critical scenarios. Till now, a large body of work has been proposed to enhance model robustness \cite{liao2018defense,athalye2018obfuscated,wong2018provable}. From a comprehensive consideration of feasibility and effectiveness, adversarial training based on projected gradient descent (PGD-AT) \cite{madry2017towards} remains one of the most promising and popular methods to improve model robustness.  Unfortunately, PGD-AT and its variants \cite{wu2020adversarial,zhang2020attacks,wu2020adversarial} substantially degrade model accuracy on benign samples, which limits their value for tasks in practice.  Remarkably, augmenting the training set with extra data can mitigate the trade-off to some extent\cite{carmon2019unlabeled}. However, additional training data is not always available due to the heavy cost of data collecting and labeling. This motivates us to explore the following question: \textbf{can we obtain both accurate and robust models at the same time without extra training data?}

\vspace{-0.8cm}
\begin{figure}[h]
        \begin{center}
                \includegraphics[scale=0.35]{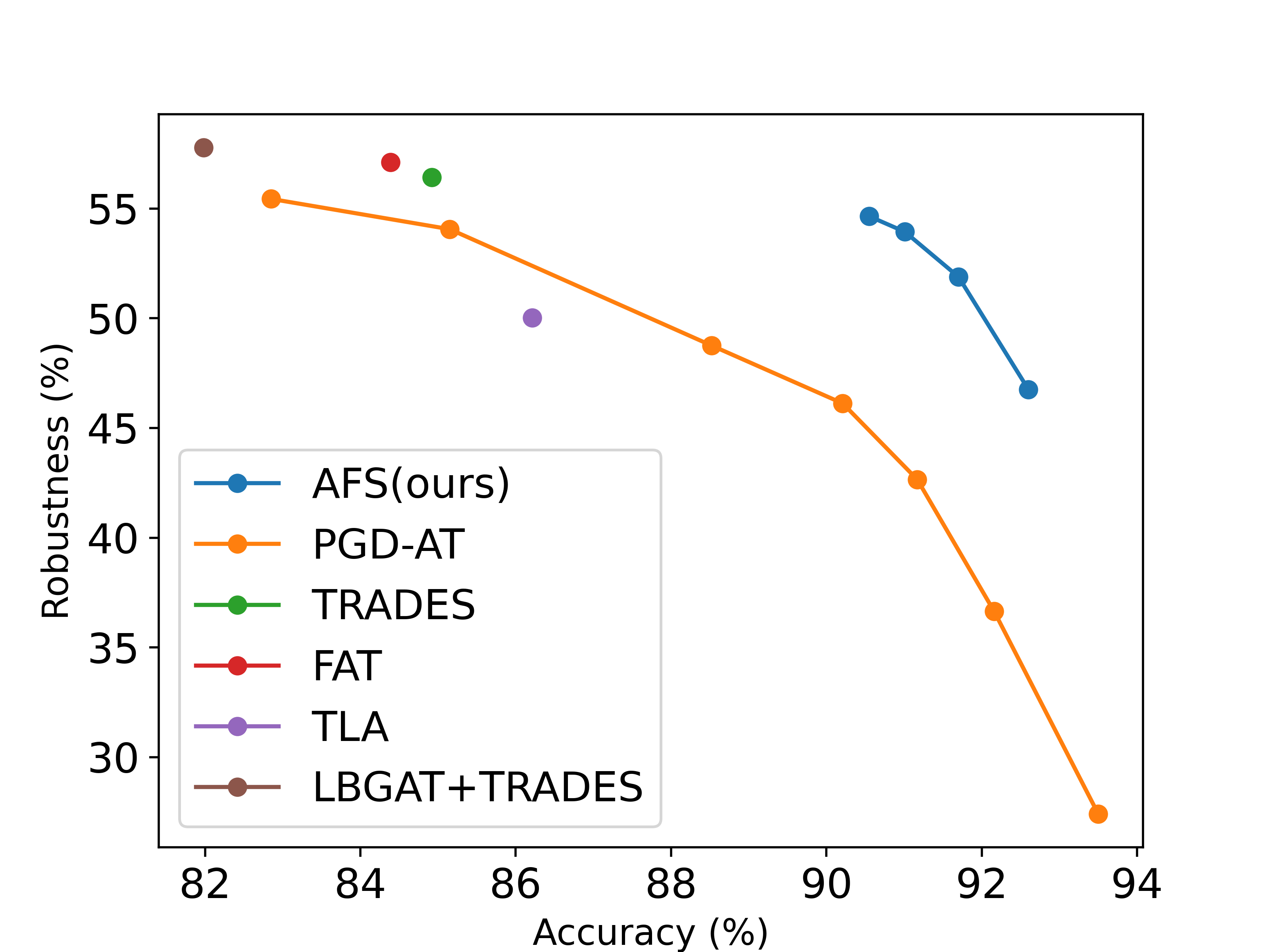}
        \end{center}
        \caption{Comparison of PGD-AT \cite{rice2020overfitting}, TRADES \cite{zhang2019theoretically}, FAT \cite{zhang2020attacks}, TLA \cite{mao2019metric}, LBGAT+TRADES \cite{cui2020learnable}, and AFS. The robustness is evaluated on CIFAR-10 using PGD attack under the perturbation budget of 8/255. Our model significantly improves the trade-off.}
        \label{comparison_robustness_accuracy}
\end{figure}
\vspace{-0.2cm}

To answer this question, we first investigate the influence of the features extracted by the networks with different training algorithms on the model accuracy and robustness. For convenience, we refer to the accurate features as the features that are highly correlated to the labels, and the robust features as the features that remain almost unchanged with adversarially perturbed inputs. Adversarial training with large perturbation budgets can hinder the network from learning non-robust but predictive features \cite{ilyas2019adversarial,tsipras2018robustness}, which are important for accurate predictions of benign samples. From this perspective, the trade-off between accuracy and robustness can be interpreted as it is difficult to train a single network to extract both robust features and non-robust but accurate features.

To solve this dilemma, we develop an adversarial feature stacking (AFS) model with a two-stage training paradigm. The AFS model combines the features extracted by multiple separately trained networks with varied levels of robustness and accuracy. Then, we adopt a linear merger to fuse the useful features to give final predictions.  Due to the availability of accurate features and robust features, the AFS model can facilitate predictions with both accuracy and robustness. We analyze the AFS model theoretically and evaluate it on CIFAR-10 and CIFAR-100 datasets with advanced adaptive attack methods, which significantly improves the trade-off between accuracy and robustness as presented in Figure \ref{comparison_robustness_accuracy}. The experimental results indicate that \textbf{it is feasible to obtain a model with high accuracy and strong robustness under limited training data}. Our key contributions are summarized as follows:

\begin{itemize}
        \item[1] A stacking model is developed to fuse the features extracted by multiple networks with different levels of robustness and accuracy. The stacking model is further analyzed theoretically and general principles are derived for selecting the basic feature extractors.
        \item[2] The AFS model is verified on CIFAR-10 and CIFAR-100 datasets with advanced attack methods. The experimental results demonstrate the feasibility to achieve both high accuracy and strong robustness without extra data.
        \item[3] We conduct extensive ablation experiments and analyze the characteristics of the linear merger, verifying the effectiveness of the AFS model.
\end{itemize}

\section{Methods}
In this section, we will introduce the overall architecture and the training methods of the proposed AFS model. In this work, we focus on the norm bounded adversarial perturbation for classification tasks. The AFS model can be generalized to other tasks and the adversarial perturbation with other constraints.
\begin{figure}[h]
        \begin{center}
                \includegraphics[scale=0.42]{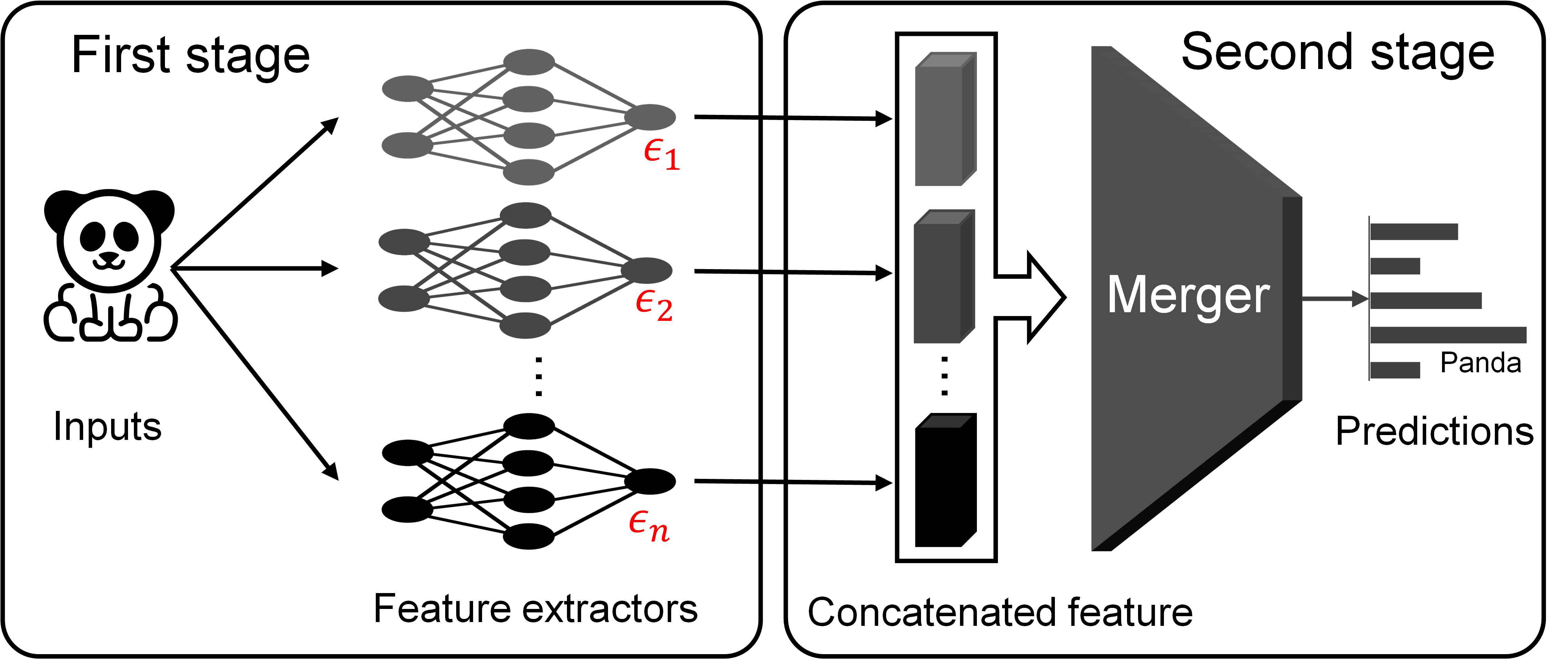}
        \end{center}
        \caption{The schematic of the AFS model. The AFS model fuses the features extracted by multiple networks with a learnable linear merger. These networks are adversarially trained with varied perturbation budgets. The classifier is optimized to select useful features for both accurate and robust predictions.}
        \label{schematic}
\end{figure}
\vspace{-0.6cm}

\subsection{Overall architecture}
The AFS model consists of two parts: several pre-trained feature extractors with varied levels of robustness and accuracy and a linear merger to fuse the features, whose overall architecture is illustrated in Figure \ref{schematic}. Generally, as discussed above, there are two types of feature extractors. For a given benign input, some of these pre-trained extractors can extract accurate features, thus contributing more accurate predictions. The others of these pre-trained extractors can extract more robust features that are not easily affected by the adversarial perturbation. The robustness and accuracy of pre-trained feature extractors can be determined by the strength of the defense methods. The processing procedure of the AFS model can be formulated as follows:
\begin{equation}
        z=w^{T} G(x;\Theta)+b,
\end{equation}
where $x$ and $z$ denote the input and the prediction, respectively. $w$ and $b$ are the trainable parameters of the linear merger. $G(x;\Theta)$ is formed by concatenating the features generated by the pre-trained extractors along the feature dimension. $\Theta$ denotes the set of parameters of all the extractors, which is presented as the following equation:
\begin{equation}
        G(x;\Theta)=\left[ g_{1}(x;\theta_{1})^T,g_{2}(x;\theta_{2})^T,...,g_{n}(x;\theta_{n})^T \right]^T,
\end{equation}
where $g_i$ is the $i$-th feature extractor parameterized by $\theta_{i}$, whose outputs are high dimensional vectors. These feature vectors are concatenated together to feed into the linear merger.

\subsection{Adversarially trained feature extractor}
In this work, we use the networks that are adversarially trained with different perturbation budgets as the feature extractors. The perturbation budgets should range from small values to large values. These networks trained with different perturbation levels can extract either more accurate features or more robust features with a large diversity. The models enhanced by other defense methods with different levels of robustness and accuracy can also be the candidates of the feature extractors. The feature extractors used in this work are separately trained with PGD-AT method \cite{madry2017towards}. The training objective of each extractor is formulated as follows:
\begin{equation}
        \min_{\theta_{i},w_i,b_i}\mathbb{E}_{(x,y)\in \mathcal{D}} \max_{\parallel \hat{x}-x \parallel_{p} \leq \epsilon_i } \mathscr{L}\left[ w_i^T g_i(\hat{x};\theta_i) +b_i,y \right],
\end{equation}
where $\epsilon_i$ is the perturbation budget for the $i$-th feature extractor and $w_i,b_i$ are the parameters of the corresponding linear classifier. $\mathcal{D}$ is the data distribution and $y$ is the true label. The inner maximization is solved by PGD method, whose iterative formula is presented as follows:
\begin{equation}
        \label{PGD_formula}
        \begin{split}
                \hat{x}_0&=x+{\rm{Uniform}}(-\epsilon_i,\epsilon_i)\\
                \hat{x}_{k+1}&=P_S\left(\hat{x}_k+\eta\cdot{\rm sgn}\left(\nabla_x\mathscr{L}\left[ w_i^T g_i(\hat{x}_k;\theta_i) +b_i,y \right]\right)\right),
        \end{split}
\end{equation}
where $\eta$ is the step size and ${\rm sgn}(\cdot)$ is the sign function. $P_S$ is an operator to project $\hat{x}_k$ into the image space. 

\subsection{Linear merger}
A naive approach to fuse the features is to take the average of the outputs of the original corresponding classifiers. However, this approach cannot adaptively select features. To address this issue, we propose a learnable linear merger to fuse the diverse features to give final predictions, which can balance the total accuracy and robustness when considering all the features. After obtaining a group of feature extractors, the linear merger is trained on the generated features. To balance the final accuracy and robustness of the whole stacking model, the features of both benign samples and adversarially perturbed samples are used to train the merger. The training objective of the linear merger is presented as follows:
\begin{equation}
        \begin{split}
                &\min_{w,b}\mathbb{E}_{(x,y)\in \mathcal{D}} \{\alpha \mathscr{L}\left[w^T G(x;\Theta) +b,y \right]      \\
                &+\max_{\parallel \hat{x}-x \parallel_{p} \leq \epsilon }(1-\alpha) \mathscr{L}\left[w^T G(\hat{x};\Theta) +b,y \right] \},
        \end{split}
\end{equation}
where $\alpha$ is the ratio ranging in $[0,1]$. $\alpha$ can balance the loss for accuracy and robustness. $\epsilon$ denotes the perturbation budget for training the merger. The crafting method of the adversarial perturbation for adversarial training is the same as the Equation \ref{PGD_formula} except for targeting the whole model. A small $\alpha$ encourages the merger to use more robust features to give robust predictions. On the contrary, a large $\alpha$ encourages the merger to use more accurate features to give accurate predictions.

\subsection{Theoretical analysis}
To understand why fusing multiple networks can improve performance, we conduct a brief theoretical analysis of the stacking model to explore the underneath mechanism. In addition, a general principle for the selection of the perturbation budgets is derived. Without loss of generality, we consider a binary classification problem with $y\in \{\pm 1\} $ and stacking three networks trained with perturbation budgets $\epsilon_1<\epsilon_2<\epsilon_3$. The error rate of each single network under the evaluation attack strength of $\Delta$ can be calculated as follows:

\begin{equation}
        \begin{split}
                {\rm{err}}_i&=\mathbb{E}_{(x,y)\in \mathcal{D}} \max_{\parallel \hat{x}-x \parallel_{p} \leq \Delta } \left[1-{\rm{sgn}}(w_i^T g_i(\hat{x};\theta_i) +b_i)\cdot y \right]/2,\\
                &=\mathbb{E}_{(x,y)\in \mathcal{D}} \max_{\parallel \hat{x}-x \parallel_{p} \leq \Delta } \left[1-{\rm{sgn}}(\hat{z}_i)\cdot y \right]/2,
        \end{split}
\end{equation}
where $\hat{z}_i$ is the logit of the network $i$ under the input perturbation:

\begin{equation}
        \hat{z}_i=w_i^T g_i(\hat{x};\theta_i) +b_i
\end{equation}

The maximum perturbation of $\hat{z}_i$, denoted by $\Delta_{z_i}$, is strongly correlated to the error rate of the network under the adversarial perturbation, which is presented as following equation:

\begin{equation}
        \Delta_{z_i}=\max_{\parallel \hat{x}-x \parallel_{p} \leq \Delta } \parallel \hat{z}_i-z_i \parallel_{p}.
\end{equation}

$\Delta_{z_i}$ decreases with training perturbation budgets, which can be validated empirically. The error rate of the stacking model with the weighted average of logits of single networks can be calculated as follows:

\begin{equation}
        {\rm{err}}=\mathbb{E}_{(x,y)\in \mathcal{D}} \max_{\parallel \hat{x}-x \parallel_{p} \leq \Delta } \left[1-{\rm{sgn}}(\Sigma_i \lambda_i \hat{z}_i)\cdot y \right]/2,
\end{equation}

where $\lambda_i>0$ and $\Sigma_i \lambda_i=1$. As an instance, to make the error rate of the stacking model lower than the single network such as the second network, a sufficient condition is to make the maximum perturbation of the logit of the stacking model smaller than that of the single network, which can be presented as the following equation:

\begin{equation}
        \Sigma_i \lambda_i \Delta_{z_i} < \Delta_{z_2}.
\end{equation}

To make the above equation hold, one solution is to make $\Delta_{z_i}$ concave with respect to the perturbation budget $\epsilon$ used to train the single network. This condition is a practical standard for the selection of perturbation budgets. Note that in the derivation, the evaluation attack strength $\Delta$ is not fixed as a specific value. Therefore, lower error rates under a range of $\Delta$ show improved trade-off. As a general case of the above logit averaging model, the proposed feature stacking model uses a learnable merger to fuse features, which is more adaptive. In practical applications, to reduce computation overhead, we can train multiple networks with evenly-spaced perturbation budgets to calculate $\Delta_{z_i}$, and then filter the candidates according to the above standard.

\section{Experiments}
In this section, we conduct a series of experiments to verify the effectiveness of the AFS model. The effects of different parameter settings on the performance are also extensively studied. Without loss of generality, we evaluate the AFS model with the adversarial perturbation bounded by infinity norm. Our code is available at \url{https://github.com/anonymous1s8f2o/afs_code}. The experimental settings are summarized as follows.

\textbf{Dataset.} We conduct experiments on widely adopted CIFAR-10 and CIFAR-100 datasets for evaluating adversarial robustness. We train feature extractors on the standard training set without extra training data.

\textbf{Network.} Unless otherwise indicated, we adopt the prevalent wide ResNet backbone with 28 layers and a width factor of 10 (WRN-28-10) \cite{zagoruyko2016wide} as the feature extractor. We use a linear classifier as the merger. 

\textbf{Training.}
The training setting of CIFAR-10 is the same as CIFAR-100. The training settings of feature extractors follow the prevalent PGD model with early stopping \cite{rice2020overfitting}. We train the linear merger 5 epochs. The perturbation budget for training the merger is set to 8/255. Other settings are the same as those of training extractors.

\textbf{Evaluation protocol.}
We evaluate the model robustness with PGD method \cite{madry2017towards}. We denote PGD-10 and PGD-20 for PGD with 10 steps and 20 steps, respectively. Stronger attack such as auto attack (AA) \cite{croce2020reliable} is also applied, which ensembles multiple gradient-based and black-box attack methods. For simplicity, we refer to the model robustness as the model accuracy on samples with a perturbation budget of 8/255. When evaluating the stacking model, the adversarial perturbation is generated for the whole model.

\subsection{Selection and analysis of basic feature extractors}

\begin{figure}[h]
        \begin{center}
                \includegraphics[scale=0.32]{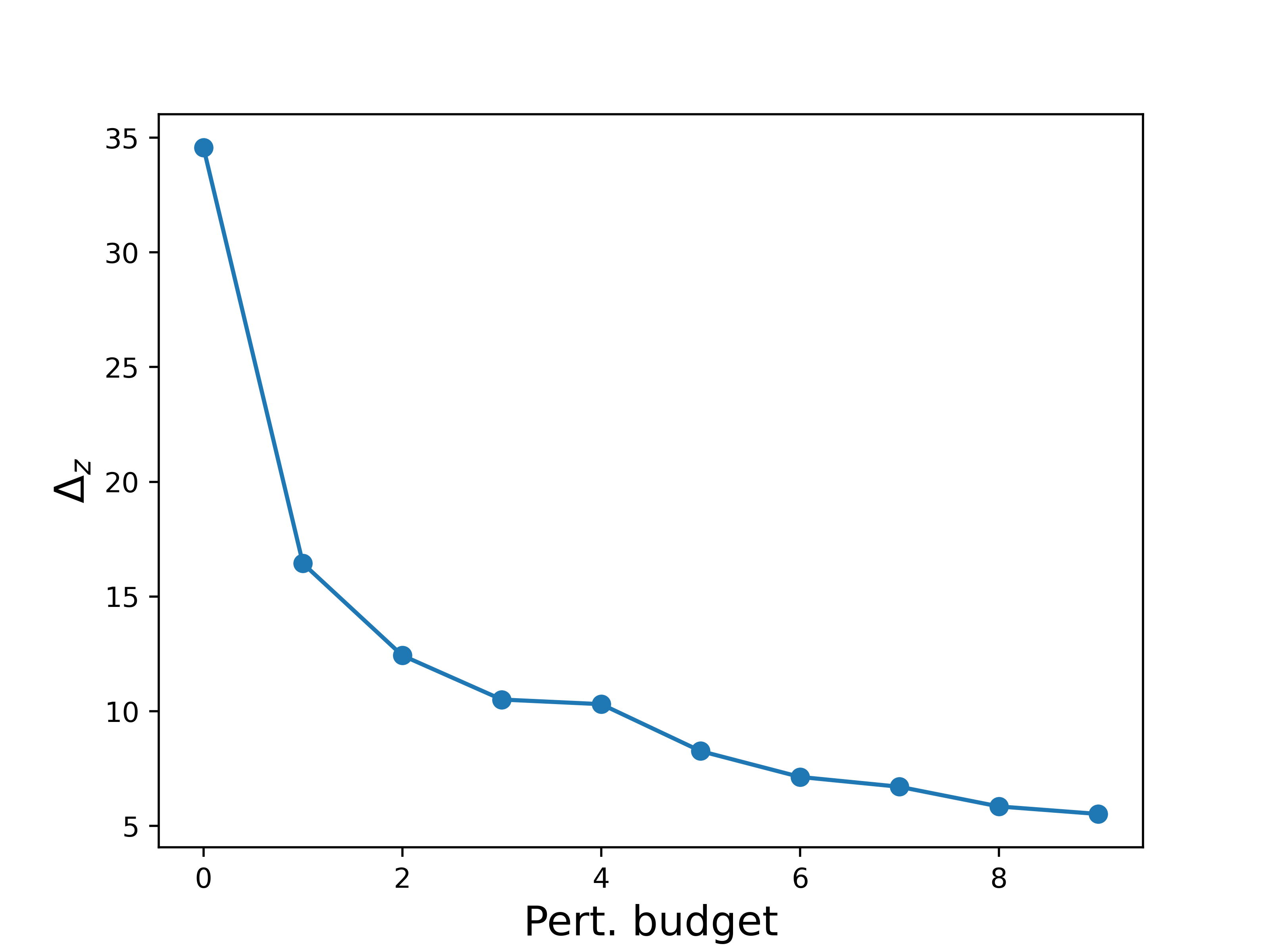}
        \end{center}
        \caption{The plot of $\Delta_{z}$ under the evaluation attack strength of 8/255 versus the training perturbation budget. $\Delta_{z_i}$ of the network trained with perturbation budgets less than 3/255 is not concave.}
        \label{deltaz}
\end{figure}

To form the stacking model, we firstly train multiple WRN-28-10 with evenly-spaced perturbation budgets from 0/255 to 9/255, which are named network-0 to network-9. The accuracy and robustness of these networks are summarized in Table \ref{extractor_evaluation}. It can be seen that the network trained with a larger perturbation budget has lower accuracy and stronger robustness. According to the selection standard, $\Delta_{z_i}$ should be concave with respect to the training perturbation. As shown in Figure \ref{deltaz}, $\Delta_{z_i}$ of the network trained with perturbation budgets less than 3/255 is not concave. Thus, these networks should not be fused. We select the networks trained with the perturbation budgets less than 9/255 and larger than 2/255 as the candidates of the feature extractors.

\vspace{-0.8cm}
\vspace{-0.8cm}

\renewcommand\tabcolsep{3.5pt}
\begin{table*}[h]
        \small
        \centering
        \caption{Accuracy and robustness of basic extractors (\%)}
        \label{extractor_evaluation}
        \begin{tabular}{c|cccccccccc}
                \toprule
                Network & 0     & 1     & 2     & 3     & 4     & 5     & 6     & 7     & 8     & 9     \\
                Pert.   & 0/255 & 1/255 & 2/255 & 3/255 & 4/255 & 5/255 & 6/255 & 7/255 & 8/255 & 9/255 \\
                \hline
                Clean   & 94.62 & 93.92 & 93.50 & 92.16 & 91.17 & 90.21 & 88.52 & 87.30 & 85.15 & 83.85 \\
                PGD-10  & 0     & 15.10 & 31.48 & 39.12 & 44.63 & 47.76 & 50.16 & 52.88 & 54.76 & 55.02 \\
                PGD-20  & 0     & 10.71 & 27.41 & 36.64 & 42.65 & 46.12 & 48.76 & 51.74 & 54.06 & 54.37 \\
                \bottomrule
        \end{tabular}
\end{table*}

\begin{figure*}[h]
        \centering
        \subfigure[]{
                \includegraphics[width=0.30\textwidth]{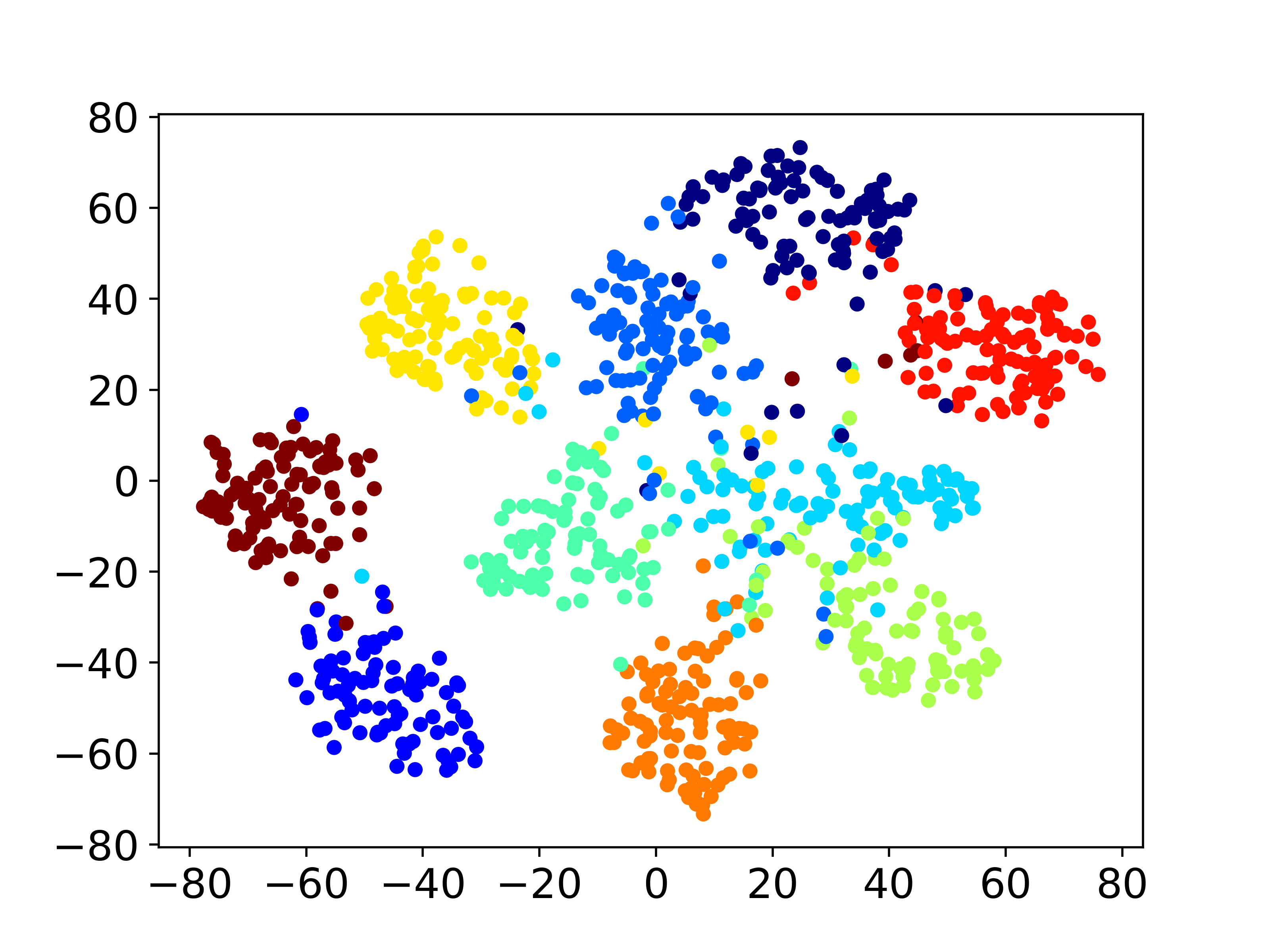}
        }
        \subfigure[]{
                \includegraphics[width=0.30\textwidth]{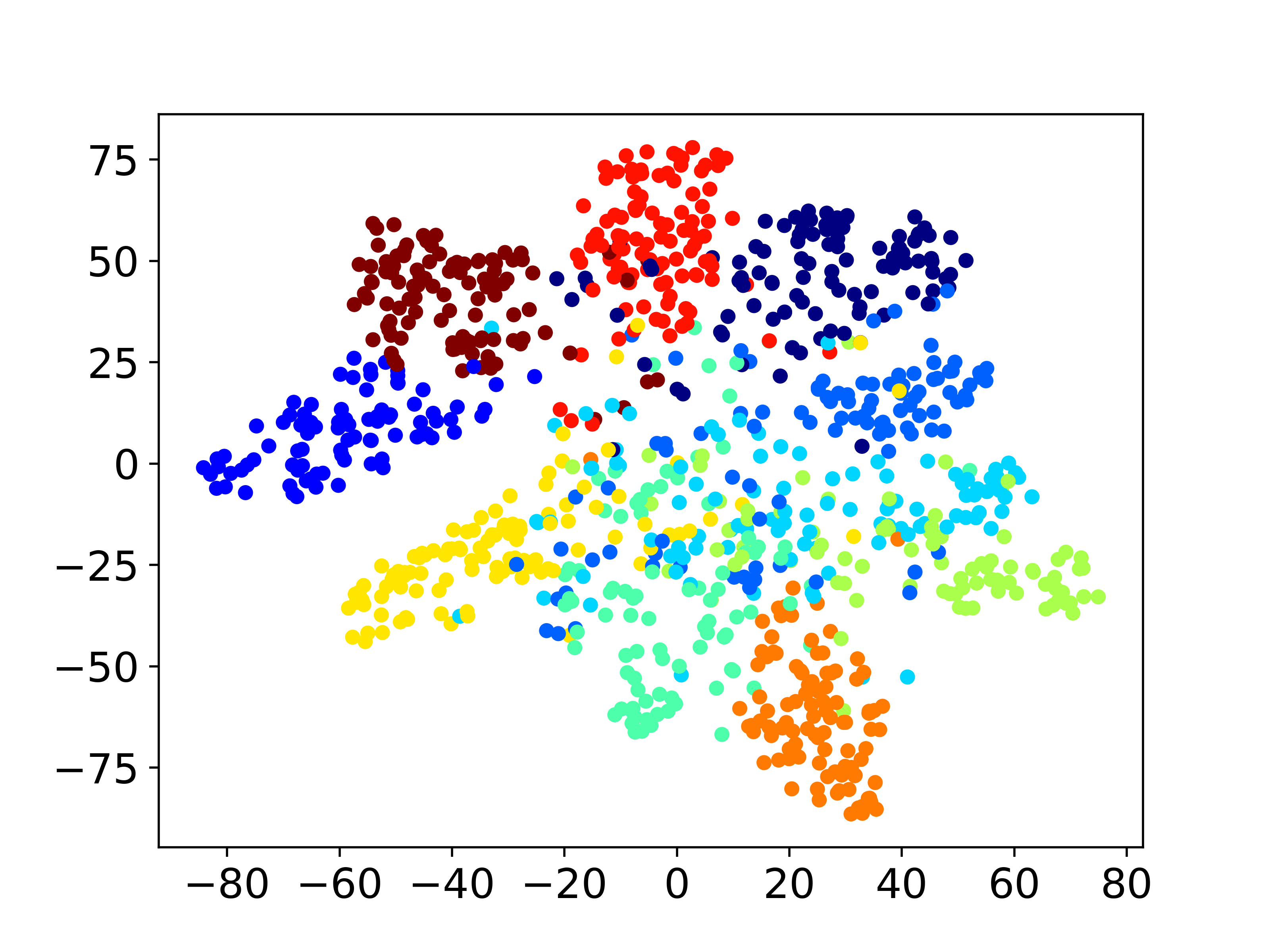}
        }
        \subfigure[]{
                \includegraphics[width=0.30\textwidth]{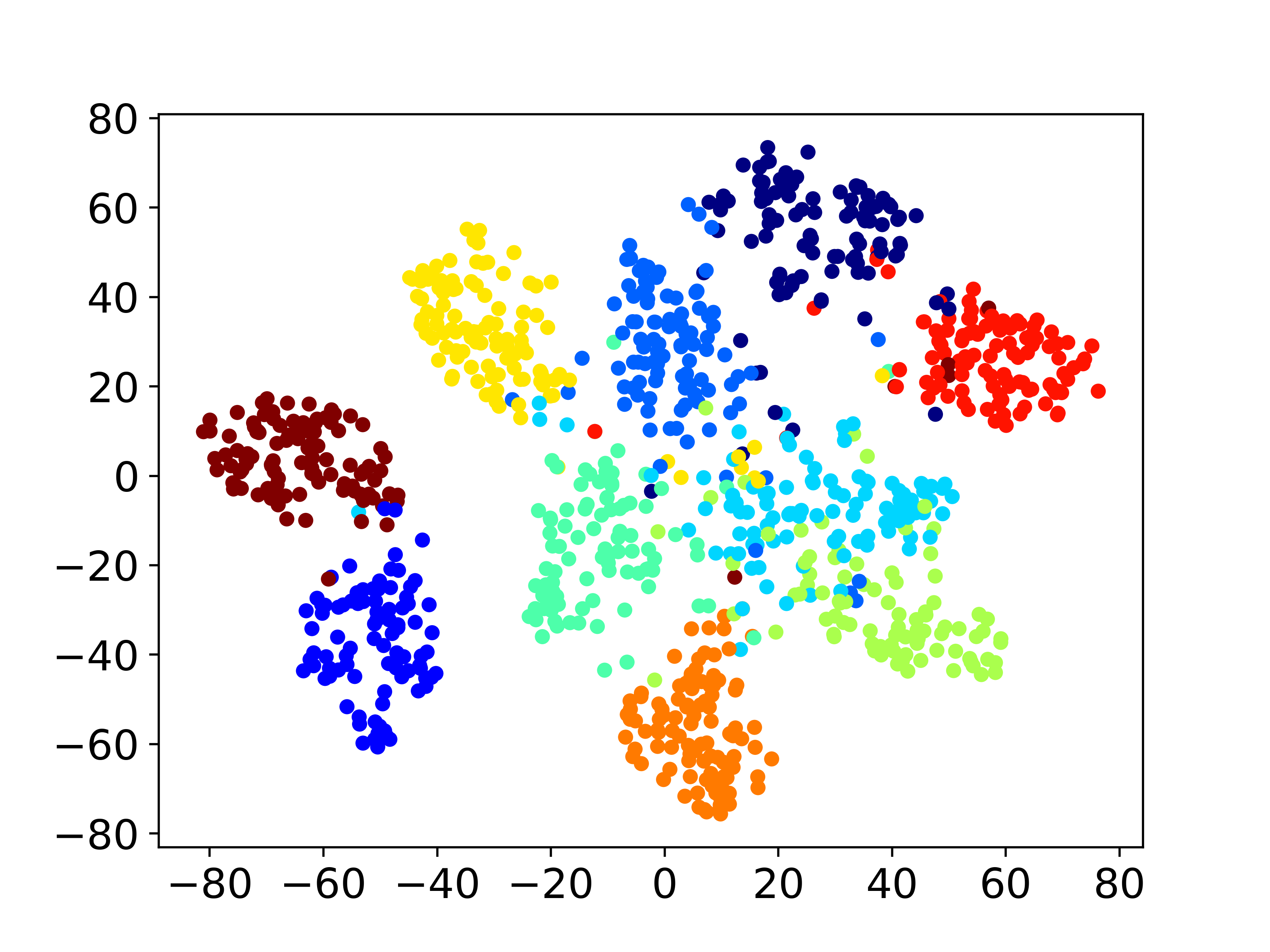}
        }
        \caption{Visualization of t-SNE embeddings. Different colors denote different classes. (a)-(c) The embeddings of the features extracted by the network-3, the network-8, and the concatenation of these two features.}
        \label{feature_tSNE}
\end{figure*}

As a preliminary analysis of the features extracted by these candidates, we visualize the 2-dimensional embeddings of some benign test samples using t-SNE method. As shown in Figure \ref{feature_tSNE}, we present the t-SNE embeddings of the features generated by the network-3 and the network-8, and the embeddings of the concatenation, respectively. The features extracted by the network trained with small perturbation budgets are separated well, but the robustness of these networks is lower. In contrast, the robustness of the network trained with large perturbation budgets is stronger, but the features extracted by these networks are not well separated, thus leading to lower accuracy on benign samples. Notably, the features formed by simply concatenating these two types of features are also separated well, indicating improved accuracy.

\vspace{-0.8cm}
\vspace{-0.6cm}

\subsection{Effect of $\alpha$ in training the merger}

\begin{figure*}[h]
        \centering
        \subfigure[]{
                \includegraphics[width=0.4\textwidth]{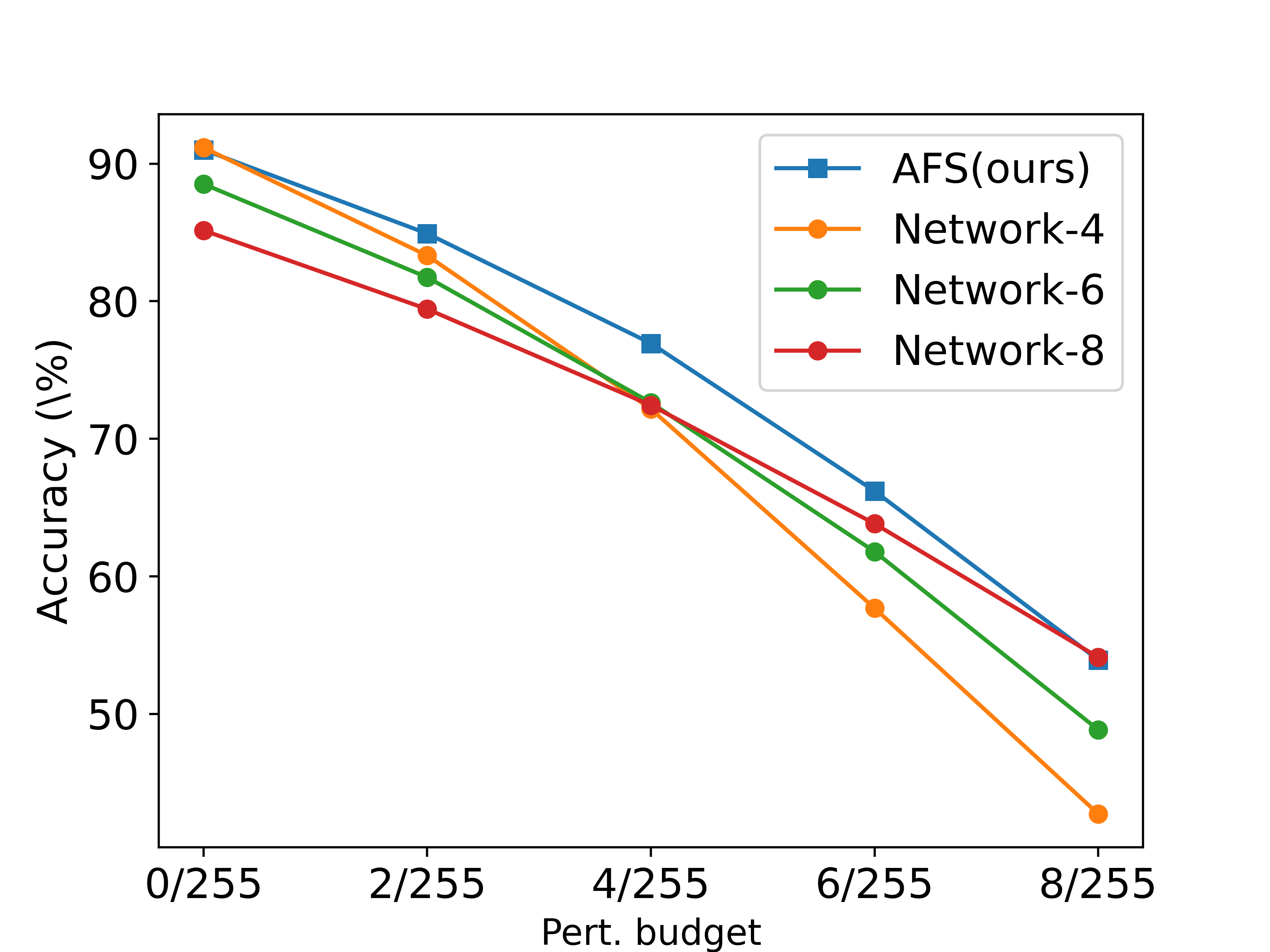}
                \label{pert_plot}
        }
        \subfigure[]{
                \includegraphics[width=0.4\textwidth]{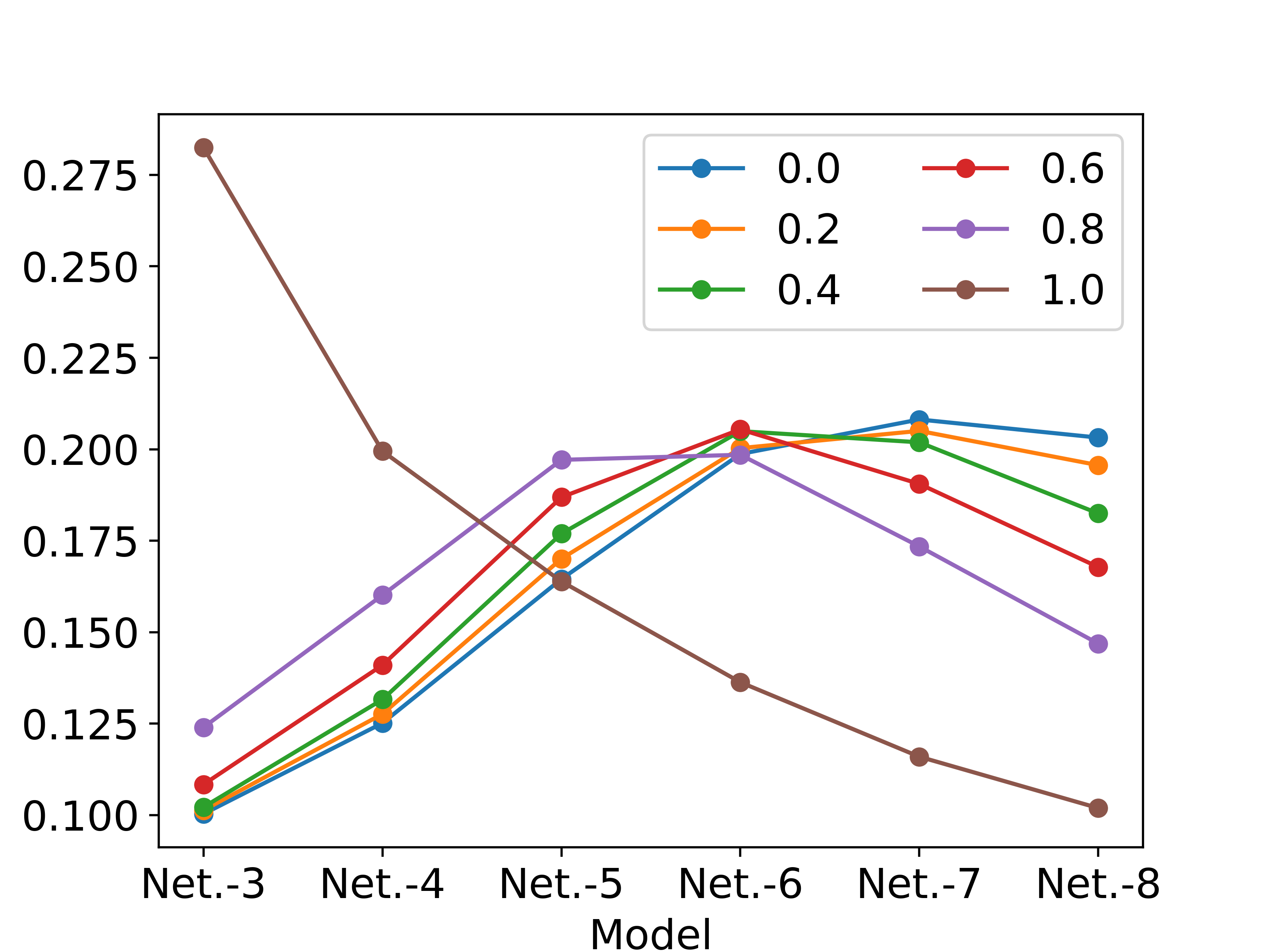}
                \label{ratio}
        }
        \caption{(a) Robustness of different models under different perturbation budgets. (b) The importance ratio of each stacked network under different $\alpha$. }
\end{figure*}

In this section, we study the effect of $\alpha$. In these experiments, we select 6 networks trained with perturbation budgets ranging from 3/255 to 8/255 as the feature extractors. The accuracy and robustness of the stacking models trained with varied $\alpha$ are summarized in Table \ref{ratio_evaluation}. The larger the $\alpha$, the higher the accuracy and the lower the robustness. However, it should be noted that the trade-off is significantly improved than single models. We evaluate the robustness of the stacking model with an $\alpha$ of 0.5 under different levels of perturbation budgets, and compare it with several single networks. The experimental results are depicted in Figure \ref{pert_plot}. The AFS model significantly improves the accuracy without sacrificing robustness. Practically, the $\alpha$ can be set to 0.5 for a better trade-off.

\begin{table}[h]
        \small
        \centering
        \caption{Accuracy and robustness of stacking models with different $\alpha$ (\%)}
        \label{ratio_evaluation}
        \renewcommand\tabcolsep{4.5pt}
        \begin{tabular}{c|cccccc}
                \toprule
                $\alpha$ & 0.0   & 0.2   & 0.4   & 0.6   & 0.8   & 1.0   \\
                \hline
                Clean    & 90.06 & 90.55 & 90.80 & 91.01 & 91.70 & 92.60 \\
                PGD-10   & 56.16 & 55.74 & 55.21 & 54.61 & 53.18 & 48.20 \\
                PGD-20   & 54.94 & 54.65 & 53.79 & 53.33 & 51.89 & 46.76 \\
                \bottomrule
        \end{tabular}
\end{table}

To investigate how the features of different extractors influence the performance, we analyze the characteristic of the weight matrix of the linear merger trained with different $\alpha$. The weight matrix can be divided into several sub-matrices. The sum of the absolute values of each sub-matrix reflects the importance of the feature generated by the corresponding extractor. Thus, we calculate the normalized sum as the importance ratio and plot it versus the stacked single model under different settings of $\alpha$ in Figure \ref{ratio}. These results demonstrate that the AFS model indeed utilizes diverse features generated by different extractors to give predictions.

\vspace{-0.8cm}
\vspace{-0.8cm}

\subsection{Effect of stacking different extractors}
In this section, we study the effect of the number of different networks to be stacked. In this experiment, we select 9 basic networks (network-0 to network-8) as the candidates to be stacked. We use a 9-dimensional binary vector to denote whether a single network is adopted. For example, '100000001' denotes the setting that network-0 and network-8 are adopted. The accuracy and robustness of different settings are summarized in Table \ref{stack_evaluation}. Generally, the more networks are stacked, the higher accuracy and the stronger robustness are. As the results show, stacking network-0, network-1 or network-2 contributes marginally to the total performance, which is consistent with the proposed selection standard. From the both consideration of performance and cost, we choose to stack the 6 networks (network-3 to network-8) as the default setting. 

\begin{table}[h]
        \small
        \centering
        \caption{Accuracy and robustness of stacking different networks (\%)}
        \label{stack_evaluation}
        \begin{tabular}{c|cccc}
                \toprule
                Setting & 100000001 & 100010001 & 101010101 & 111111111 \\
                \hline
                Clean   & 86.25     & 90.00     & 90.79     & 91.08     \\
                PGD-10  & 53.30     & 53.01     & 53.77     & 54.90     \\
                PGD-20  & 52.44     & 51.96     & 52.44     & 53.38     \\
                \bottomrule
        \end{tabular}
\end{table}
\vspace{-0.4cm}

\subsection{Comparison with state-of-the-art methods}
The default AFS model is evaluated on CIFAR-10 and CIFAR-100 with advanced attacks such as AA, and compared with state-of-the-art defense methods, whose results are summarized in Table \ref{comparison_cifar10} and Table \ref{comparison_cifar100}. The trade-off is measured by the mean of the clean accuracy and the robustness evaluated by AA. The results listed at the bottom part of the table are our implementations. Other results are cited from original papers. All the methods are evaluated with the WRN-34-10 network architecture. The AFS model in this work is based on PGD-AT. Compared to the PGD-AT with a perturbation budget of 8/255, our AFS model achieves a benign accuracy improvement of $\sim$6\% on CIFAR-10 and $\sim$10\% on CIFAR-100 with stronger robustness. These results indicate that we can obtain a model with both high accuracy and strong robustness without extra training data. Additionally, as shown in Figure \ref{comparison_robustness_accuracy}, the AFS model outperforms other methods in terms of the trade-off.

\vspace{-0.4cm}
\begin{table}[h]
        \small
        \centering
        \caption{Comparison of different methods on CIFAR-10 (\%)}
        \label{comparison_cifar10}
        \begin{tabular}{c|ccc|c}
                \toprule
                Method                               & Clean          & PGD            & AA             & Trade-off      \\
                \hline
                FAT \cite{zhang2020attacks}          & 84.39          & 57.12          & \textbf{53.51} & 68.95          \\
                TLA \cite{mao2019metric}             & 86.21          & 50.03          & 47.41          & 66.81          \\
                TRADES \cite{zhang2019theoretically} & 84.92          & 56.43          & 53.08          & 69             \\
                LBGAT+TRADES \cite{cui2020learnable} & 81.98          & \textbf{57.78} & 53.14          & 67.56          \\
                \hline
                PGD-AT  \cite{rice2020overfitting}   & 85.77          & 55.53          & 52.08          & 68.92          \\
                AFS(ours)                            & \textbf{90.93} & 54.70          & 53.05          & \textbf{71.99} \\
                \bottomrule
        \end{tabular}
\end{table}
\vspace{-0.8cm}

\vspace{-0.4cm}
\begin{table}[h]
        \small
        \centering
        \caption{Comparison of different methods on CIFAR-100 (\%)}
        \label{comparison_cifar100}
        \begin{tabular}{c|ccc|c}
                \toprule
                Method                               & Clean          & PGD            & AA             & Trade-off      \\
                \hline
                LBGAT+TRADES \cite{cui2020learnable} & 60.43          & \textbf{35.50} & \textbf{29.34} & 44.88          \\
                \hline
                PGD-AT  \cite{rice2020overfitting}   & 60.68          & 30.45          & 26.59          & 43.63          \\
                AFS(ours)                            & \textbf{70.54} & 29.03          & 27.36          & \textbf{48.95} \\
                \bottomrule
        \end{tabular}
\end{table}
\vspace{-0.8cm}

\subsection{Ablation study}
To investigate the key points that make AFS model work, we conduct several ablation experiments on CIFAR-10. The parameters, computational costs, and performance, are presented in Table \ref{ablation_study}. The baseline model is a WRN-28-10 trained with a perturbation budget of 8/255. For comparison, we train a single WRN-28-10 with random perturbation budgets ranging from 3/255 to 8/255. The accuracy of this model improves about 4\% at the cost of the same level robustness degradation, thus failing to improve the trade-off between accuracy and robustness than the baseline model. The results indicate that training a single network with different perturbation budgets cannot alleviate the trade-off.

To investigate the effect of the model capacity on the trade-off, we train a large WRN-28-30 model, whose depth is the same as the default AFS model. With large width, the parameters and computational costs of the large model are 1.5 times larger than the default AFS model. The accuracy of this large model improves about 2.5\% without degrading robustness. The results demonstrate that increasing model capacity can improve the trade-off within certain limits but the improvement is not remarkable. In contrast, with even fewer parameters and smaller computational costs than the above mentioned large model, the AFS model can significantly improve the accuracy about 6\% without degrading robustness.

\begin{table}[h]
        \small
        \centering
        \caption{Comparison of different ablation settings}
        \label{ablation_study}
        \begin{tabular}{c|cc|cc}
                \toprule
                Setting     & Para. (M) & MACs (G) & Clean & PGD-20 \\
                \hline
                Baseline    & 36.47    & 5.24     & 85.15 & 54.06  \\
                Rand.pert.  & 36.47    & 5.24     & 89.02 & 49.73  \\
                Large model & 328      & 47.04    & 87.67 & 54.08  \\
                AFS         & 218.82   & 31.44    & 91.01 & 53.94  \\
                \bottomrule
        \end{tabular}
\end{table}
\vspace{-0.8cm}
\vspace{-0.4cm}
\subsection{Weight analysis}
\begin{figure*}[h]
        \centering
        \subfigure[]{
                \includegraphics[width=0.29\textwidth]{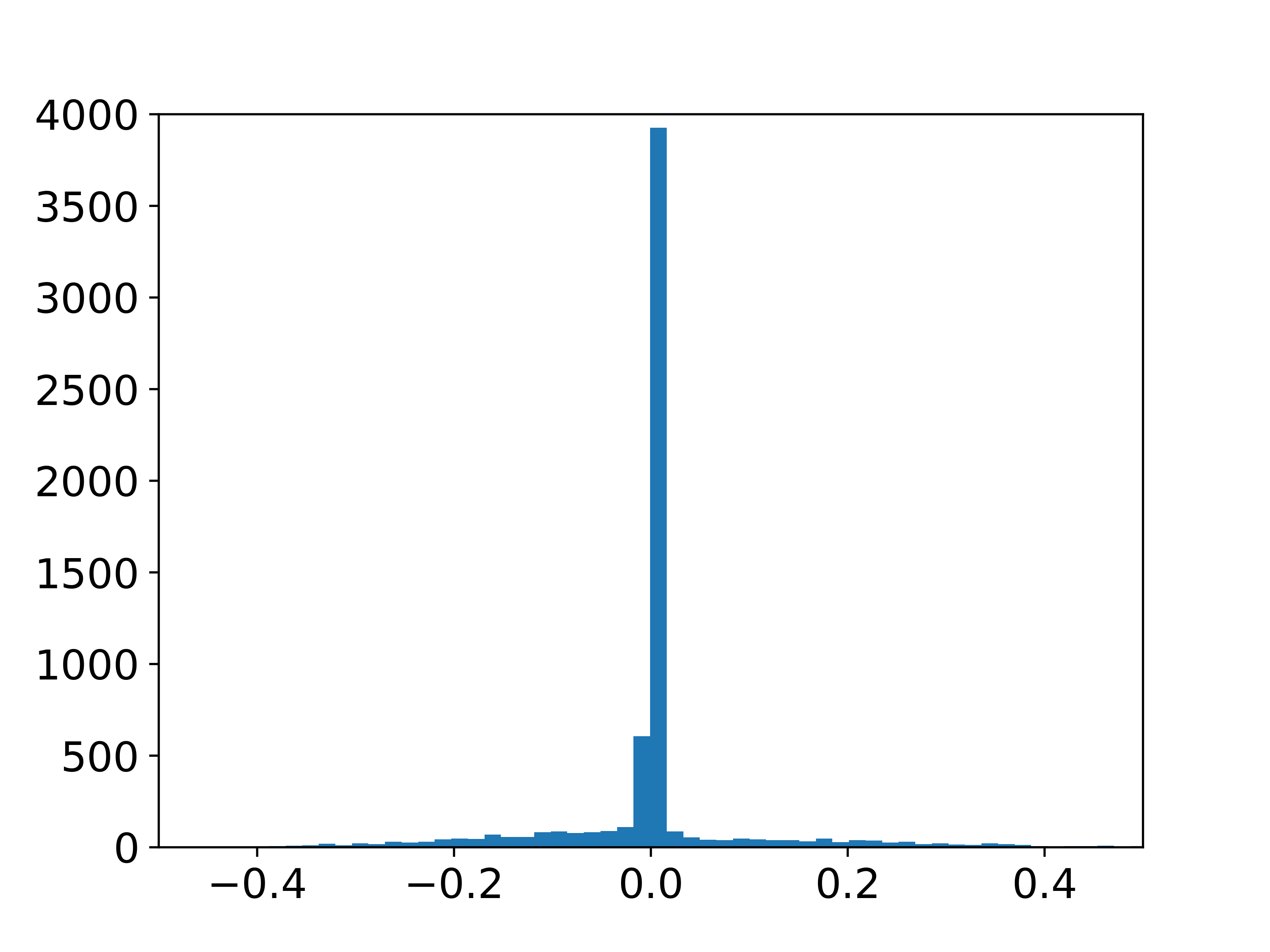}
                \label{histogram_weight_1}
        }
        \subfigure[]{
                \includegraphics[width=0.29\textwidth]{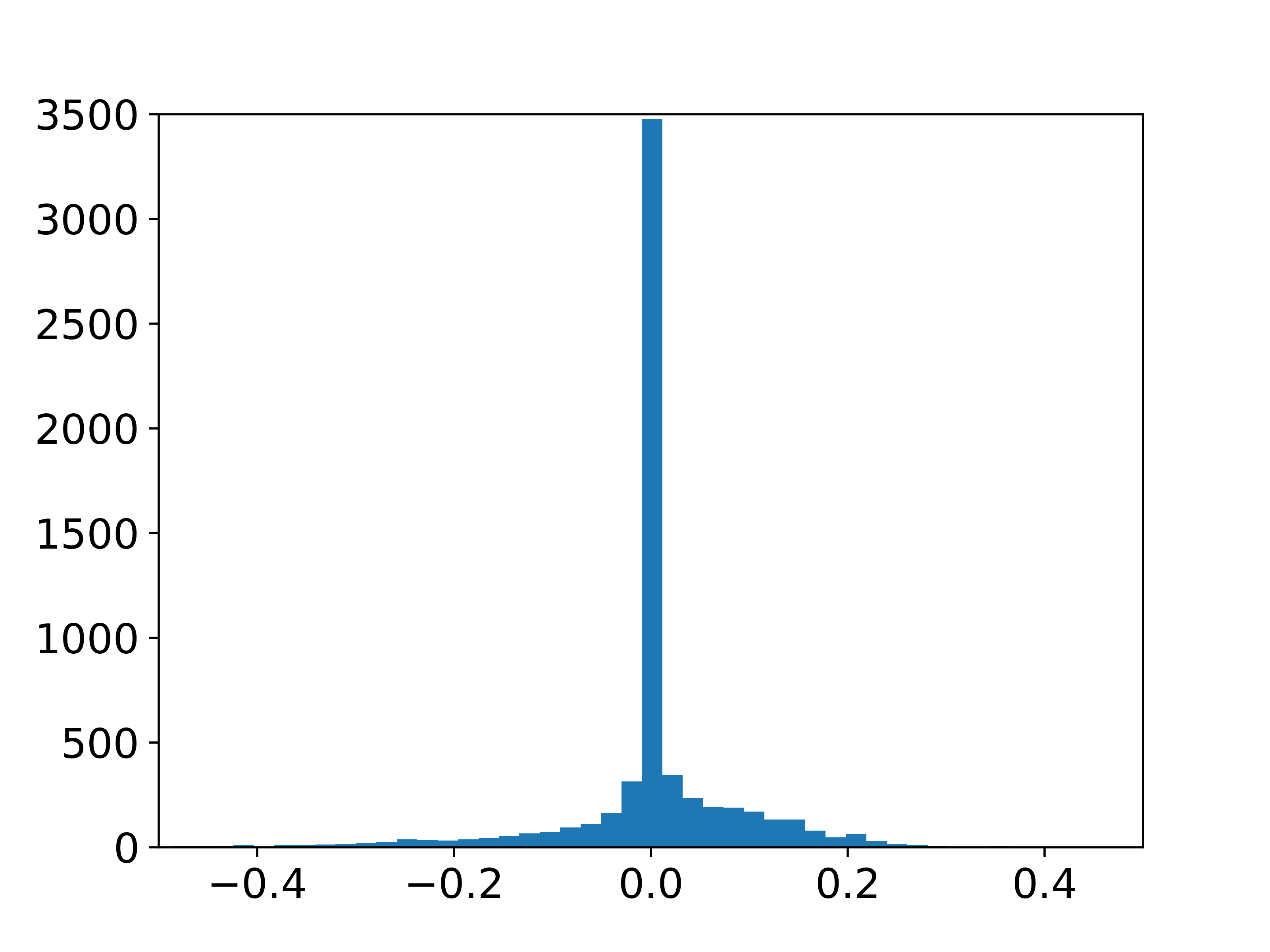}
                \label{histogram_weight_8}
        }
        \subfigure[]{
                \includegraphics[width=0.29\textwidth]{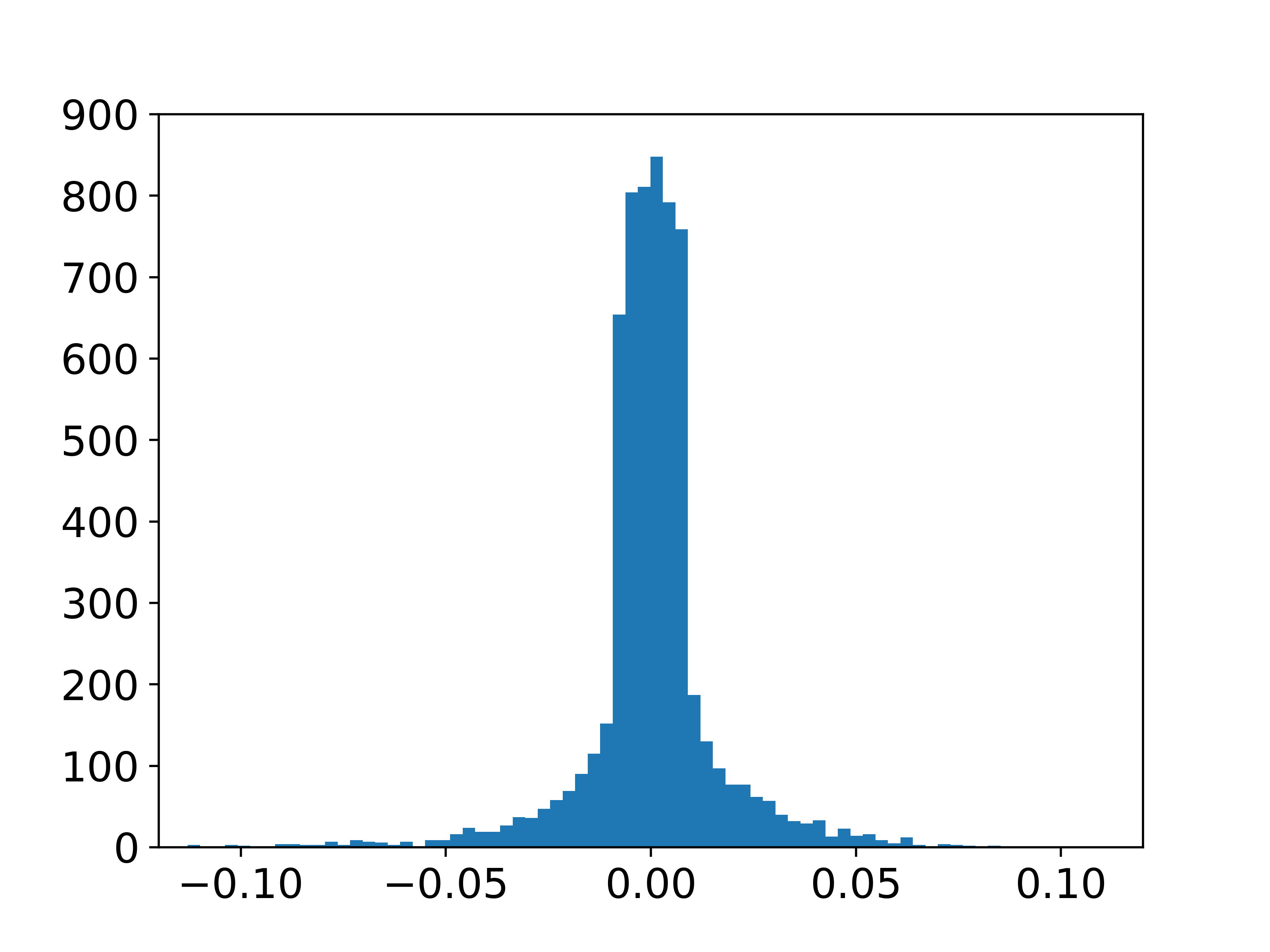}
                \label{histogram_weight_345678}
        }
        \caption{ Histograms of the weight matrix of linear classifiers. (a)-(c) The histogram of the network-1, the network-8, and the AFS model with the default setting.}
\end{figure*}

 We present the histogram of the weight matrix of the AFS model with the default setting (Figure \ref{histogram_weight_345678}), and compare it with that of the weight matrix of the classifier of the network-1 and the network-8 (Figure \ref{histogram_weight_1},\ref{histogram_weight_8}). The weight matrices of the network-1 and the network-8 are relatively sparse and have a small part of larger values, indicting that the corresponding classifier heavily depends on a few key features to give predictions. In contrast, the weight matrix of the linear merger of the AFS model is distributed more evenly, demonstrating that the AFS model uses more diverse features.
 
\vspace{-0.8cm}

\vspace{-0.6cm}
\vspace{-0.6cm}
\section{Conclusion}
In this work, we propose the AFS model that deploys a linear merger to fuse the features extracted by multiple networks adversarially pre-trained with different levels of perturbation budgets. The AFS model is verified on CIFAR-10 and CIFAR-100 datasets with advanced attack methods, which significantly outperforms the state-of-the-art methods. The experimental results demonstrate the feasibility to obtain models with both high accuracy and strong robustness without extra training data. In the future, the AFS model is expected to be combined with more advanced defense methods and trained on more data to further improve the accuracy and robustness.
\section{Acknowledgements}
This work was partly supported by the National Key Research and Development Program of China (No. 2021ZD0200300) and the National Nature Science Foundation of China (No. 61836004).
\bibliographystyle{splncs04}
\bibliography{sn-bibliography}

\begin{thebibliography}{10}
\providecommand{\url}[1]{\texttt{#1}}
\providecommand{\urlprefix}{URL }
\providecommand{\doi}[1]{https://doi.org/#1}

\bibitem{athalye2018obfuscated}
Athalye, A., Carlini, N., Wagner, D.: Obfuscated gradients give a false sense
  of security: Circumventing defenses to adversarial examples pp. 274--283
  (2018)

\bibitem{carmon2019unlabeled}
Carmon, Y., Raghunathan, A., Schmidt, L., Duchi, J.C., Liang, P.S.: Unlabeled
  data improves adversarial robustness. In: Advances in Neural Information
  Processing Systems. pp. 11192--11203 (2019)

\bibitem{croce2020reliable}
Croce, F., Hein, M.: Reliable evaluation of adversarial robustness with an
  ensemble of diverse parameter-free attacks pp. 2206--2216 (2020)

\bibitem{cui2020learnable}
Cui, J., Liu, S., Wang, L., Jia, J.: Learnable boundary guided adversarial
  training pp. 15721--15730 (2021)

\bibitem{ilyas2019adversarial}
Ilyas, A., Santurkar, S., Tsipras, D., Engstrom, L., Tran, B., Madry, A.:
  Adversarial examples are not bugs, they are features. In: Advances in Neural
  Information Processing Systems. pp. 125--136 (2019)

\bibitem{lecun2015deep}
{LeCun}, Y., {Bengio}, Y., {Hinton}, G.: Deep learning. Nature
  \textbf{521}(7553),  436--444 (2015)

\bibitem{li2019super}
Li, H., Li, G., Shi, L.: Super-resolution of spatiotemporal event-stream image.
  Neurocomputing  \textbf{335},  206--214 (2019)

\bibitem{liao2018defense}
Liao, F., Liang, M., Dong, Y., Pang, T., Hu, X., Zhu, J.: Defense against
  adversarial attacks using high-level representation guided denoiser. In:
  Proceedings of the IEEE Conference on Computer Vision and Pattern
  Recognition. pp. 1778--1787 (2018)

\bibitem{liu2021adversarial}
Liu, F., Xu, M., Li, G., Pei, J., Shi, L., Zhao, R.: Adversarial symmetric
  gans: Bridging adversarial samples and adversarial networks. Neural Networks
  \textbf{133},  148--156 (2021)

\bibitem{madry2017towards}
Madry, A., Makelov, A., Schmidt, L., Tsipras, D., Vladu, A.: Towards deep
  learning models resistant to adversarial attacks. arXiv preprint
  arXiv:1706.06083  (2017)

\bibitem{mao2019metric}
Mao, C., Zhong, Z., Yang, J., Vondrick, C., Ray, B.: Metric learning for
  adversarial robustness. In: Advances in Neural Information Processing
  Systems. pp. 480--491 (2019)

\bibitem{pei2019towards}
Pei, J., Deng, L., Song, S., Zhao, M., Zhang, Y., Wu, S., Wang, G., Zou, Z.,
  Wu, Z., He, W., et~al.: Towards artificial general intelligence with hybrid
  tianjic chip architecture. Nature  \textbf{572}(7767),  106--111 (2019)

\bibitem{rice2020overfitting}
Rice, L., Wong, E., Kolter, Z.: Overfitting in adversarially robust deep
  learning. In: International Conference on Machine Learning. pp. 8093--8104.
  PMLR (2020)

\bibitem{szegedy2014intriguing}
{Szegedy}, C., {Zaremba}, W., {Sutskever}, I., {Bruna}, J., {Erhan}, D.,
  {Goodfellow}, I., {Fergus}, R.: Intriguing properties of neural networks. In:
  ICLR 2014 : International Conference on Learning Representations (ICLR) 2014
  (2014)

\bibitem{tsipras2018robustness}
Tsipras, D., Santurkar, S., Engstrom, L., Turner, A., Madry, A.: Robustness may
  be at odds with accuracy  (2018)

\bibitem{wong2018provable}
Wong, E., Kolter, Z.: Provable defenses against adversarial examples via the
  convex outer adversarial polytope. In: International Conference on Machine
  Learning. pp. 5286--5295. PMLR (2018)

\bibitem{wu2020adversarial}
Wu, D., Xia, S.T., Wang, Y.: Adversarial weight perturbation helps robust
  generalization. Advances in Neural Information Processing Systems
  \textbf{33} (2020)

\bibitem{zagoruyko2016wide}
Zagoruyko, S., Komodakis, N.: Wide residual networks. arXiv preprint
  arXiv:1605.07146  (2016)

\bibitem{zhang2019theoretically}
Zhang, H., Yu, Y., Jiao, J., Xing, E., El~Ghaoui, L., Jordan, M.: Theoretically
  principled trade-off between robustness and accuracy pp. 7472--7482 (2019)

\bibitem{zhang2020attacks}
Zhang, J., Xu, X., Han, B., Niu, G., Cui, L., Sugiyama, M., Kankanhalli, M.:
  Attacks which do not kill training make adversarial learning stronger pp.
  11278--11287 (2020)

\end{thebibliography}
\end{document}